\documentclass{article}

\usepackage{arxiv}

\usepackage[utf8]{inputenc} 
\usepackage[T1]{fontenc}    
\usepackage{hyperref}       
\usepackage{url}            
\usepackage{booktabs}       
\usepackage{amsfonts}       
\usepackage{nicefrac}       
\usepackage{microtype}      
\usepackage{lipsum}
\usepackage{graphicx}
\graphicspath{ {./img/} }

\usepackage{cite}
\usepackage{graphicx}
\usepackage{textcomp}
\usepackage{xcolor}
\usepackage[ruled,vlined,linesnumbered,noend]{algorithm2e}
\usepackage{amsmath}
\usepackage[bottom]{footmisc}

\DeclareMathOperator{\TrainModel}{train\_model}
\DeclareMathOperator{\EvaluateModel}{evaluate}
\DeclareMathOperator{\Len}{len}
\DeclareMathOperator{\Min}{min}
\DeclareMathOperator{\Max}{max}
\DeclareMathOperator{\Randint}{randint}
\DeclareMathOperator{\RandomChoice}{random.choice}
\DeclareMathOperator{\RandomRandom}{random.random}
\DeclareMathOperator{\DefaultDict}{defaultdict}
\DeclareMathOperator{\ReturnV}{return}

\title{Ensemble neuroevolution based approach for multivariate time series anomaly detection}

\author{
Kamil Faber\\
Department of of Computer Science\\
  AGH University of Science and Technology\\
  Cracow, Poland \\
  \texttt{kfaber@agh.edu.pl} \\
 \And
 Dominik \.{Z}urek \\
  Department of Computer Science\\
  AGH University of Science and Technology\\
  Cracow, Poland \\
  \texttt{dzurek@agh.edu.pl} \\
  \And
Marcin Pietro\'n \\
  Department of of Computer Science\\
  AGH University of Science and Technology\\
  Cracow, Poland \\
  \texttt{pietron@agh.edu.pl} \\
   \And
   Kamil Pi\k{e}tak \\
  Department of of Computer Science\\
  AGH University of Science and Technology\\
  Cracow, Poland \\
  \texttt{kpietak@agh.edu.pl} \\

}

\begin{document}
\maketitle

\begin{abstract}
Multivariate time series anomaly detection is a very common problem in the field of failure prevention. Fast prevention means lower repair costs and losses. The amount of sensors in novel industry systems makes the anomaly detection process quite difficult for humans. Algorithms which automates the process of detecting anomalies are crucial in modern failure-prevention systems. Therefore, many machine and deep learning models have been designed to address this problem. Mostly, they are autoencoder-based architectures with some generative adversarial elements. In this work, a framework is shown which incorporates neuroevolution methods to boost the anomaly-detection scores of new and already known models. The presented approach adapts evolution strategies for evolving ensemble model, in which every single model works on a subgroup of data sensors. The next goal of neuroevolution is to optimise architecture and hyperparameters like window size, the number of layers, layer depths, etc. The proposed framework shows that it is possible to boost most of the anomaly detection deep learning models in a reasonable time and a fully automated mode. The tests were run on SWAT and WADI datasets. To our knowledge, this is the first approach in which an ensemble deep learning anomaly detection model is built in a fully automatic way using a neuroevolution strategy.
\end{abstract}


\section{Introduction}
In the paper, we propose a high-level ensemble approach which is fine tuned by a neuroevolution algorithm. The presented method is model independent. It can be adapted to any deep learning anomaly detection model. The main advantage of the algorithm is its fully automated mode. 

In the anomaly detection field, the deep learning models    are those which achieve the best results on well-known benchmarks. These are mainly deep autoencoders based on LSTM layers, convolutional or fully connected sequence  of layers. A wide variety of autoencoders are used, such as variational, denoise or adversarial autoencoders. Research shows that some further improvements like adding a discriminator as an additional verification module or some other GAN based autoencoder modifications can boost detection results. Recently we can also observe promising results in using deep graph neural networks in anomaly detection \cite{graph-nn}.

Neuroevolution is a form of artificial intelligence that uses evolutionary algorithms to generate artificial neural networks (ANN), parameters, topology and rules. The most popular algorithms are NEAT, HyperNEAT, coDeepNEAT etc.
In the presented approach, is partially based on NEAT algorithm which is used for generating an optimal anomaly detection model. The search space and crossover/mutation rules are defined. The novelty of the proposed algorithm is that new search dimensions have been added. These dimensions are training data distribution, dividing data to subgroups and searching for the optimal composition of the ensemble model. 

The proposed neuroevolution search space is based on forming encoders and decoders from single neural layers like fully connected, convolutional, recurrent or attention layers. There are two main dimensions of optimisation. Therefore, two populations are inside the algorithm. The first is the models population from which new single models are evolved by genetic operators. 
The second is the subgroup population, which is needed to form the ensemble model from the models population. This work concentrates on the data optimisation stage and the setting up of the ensemble model. It shows how this aspect can improve non-ensemble models. The last step in the NAS (neural architecture search) is fitness definition. In the presented approach, the fitness is the sum of F1 scores from the training dataset and from the random reduced validation dataset.

The main advantages of the presented algorithm are that enables building the ensemble model in automatic mode and creates a wide search space between various deep learning autoencoders, GAN architectures and optimal training data subgroups. 

\section{Related works}
Anomaly detection has recently become quite a popular research subject. The basic unsupervised methods include linear model-based methods \cite{PCA}, distance-based methods \cite{KNN}\cite{LOF}, density-based methods \cite{one-class-svm}, isolation based-methods \cite{Isolation_forest} and many others. The best f1-score for these methods is ~23\% on SWAT and 9\% on WADI datasets. However, deep learning-based methods have recently gained significant improvements in anomaly detection over the aforementioned approaches. One of the most popular deep learning models for multivariate anomaly detection are auto-encoder models (AE), which use the reconstruction error as an anomaly inspection. Zong \textit{et al.} proposes a deep autoencoding Gaussian  mixture model (DAGMM) \cite{DAGMAM} which jointly optimises the parameters of the deep autoencoder and the mixture model simultaneously. This solution yields an f1-score of 55\% for SWAT and 20\% for WADI datasets. Park \textit{et al.} introduced the LSTM-VAE model \cite{lstm-vae} which replaces the feed-forward network in the variational autoencoder (VAE) with LSTM. As a result of this approach, it was possible to gain an f1-score of 75\% for SWAT and 25\% for WADI datasets. Russo  \textit{et al.} use an autoencoder which consists of 1D convolution layers \cite{cnn_auto-encoder}. This model was tested with  \textit{the Urban Water Observatory Initiative (www.eawag.ch/uwo)} datasets and has an anomaly detection accuracy of 35\%. Audibert \textit{et al.} proposed a fast and stable method called USAD \cite{USAD} which is based on adversely trained autoencoders. This model contains only fully connected layers and achieves a 79\% detection anomaly for SWAT and 23\% for WADI dataset. Generative adversarial networks (GANs) as anomaly detectors were proposed in \cite{MAD-GAN}. The authors used LSTM as the generator and discriminator models in the GAN framework and anomalies were detected by the use of a combination of both model errors (DR-Score). Through the use of this approach, the anomaly accuracy for this model is 77\% for SWAT and 37\% for WADI datasets. Deng \textit{et al.} \cite{graph-nn} achieves an f1-score of 81\% for SWAT and 57\% for WADI datasets through the use of a graph neural network (GNN). The mentioned deep learning models - LSTM, USAD and CNN 1D are the baseline for proposed in this paper solutions.
Recently neuroevolution algorithms are used in many machine learning tasks for improving accuracy for deep learning models \cite{neuroevol_overview}.  In \cite{sceneNet} the neuroevolution search is used for evolving neural networks for object classification in high resolution remote sensing images. In \cite{neuro_image} authors present a neuroevolution algorithm for standard image classification. Authors in \cite{neuro_evolving} show the neuroevolution strategy scheme for language modeling, image classification and object detection tasks. It is based on the co-evolutionary NEAT algorithm which has two levels of optimization. The first one is single deep learning sub-block optimization. The second one is composition of sub-blocks to form a whole network. Presented results showed that in most of the cases optimized models achieved better results than models designed by humans.    

\section{Autoencoder architecture}
Autoencoders are an unsupervised learning technique in which the neural network is trained to learn the
compressed representation of raw data. The model consists of two parts: encoder \textit{E} and decoder \textit{D}. The encoder learns how to efficiently compress and encode the input data \textit{X} to represent them in reduced dimensionality - latent variables \textit{Z}. The decoder is taught how to reconstruct the latent variables \textit{Z} back to its original shape. The model is trained to minimise \textit{reconstruction loss} which means reducing the difference between the output of the decoder and the original input data, which can be expressed as:
\begin{equation}
    \mathcal{L}(X, \hat{X}) = ||X - AE(X)||_2
\end{equation}
where 
\begin{equation}
    AE(X) = D(Z),   Z = E(X)
\end{equation}
The simplest kind of autoencoder is an \textit{Undercomplete autoencoder (UAE)}. These models  learn the most important and relevant attributes of the input data through the use of a bottleneck with a smaller dimension than the input data. Another type of autoencoder, called the \textit{Denoie autoencoder (DAE)}, extracts important features from the data through reconstructing the original input after it has been contaminated by noise. In unsupervised tasks, the most popular type of autoencoders are \textit{variational autoencoders (VAE)}. These autoencoders replace the bottleneck vector with two vectors: one for representing the mean of the distribution and the second for representing the standard deviation of the distribution. VAE's for a given input in the encoding phase determine a distribution of the latent variables. By contrast the decoder determines the distribution of the inputs corresponding to the given latent variables.

Autoencoders are widely used in many fields, such as online intrusion detection \cite{Mirsky2018KitsuneAE}, malware detection \cite{2020MalwareDetection} and anomaly detection in streaming data \cite{2018StreamLearning}. This kind of model can consist of various types of layers e.g. fully connected layers, CNN, LSTM etc.

\section{Neuroevolution ensemble approach}
\label{neuro-evolution}

\begin{figure*}[ht]
  \centering\includegraphics[width=0.75\linewidth,clip=true]{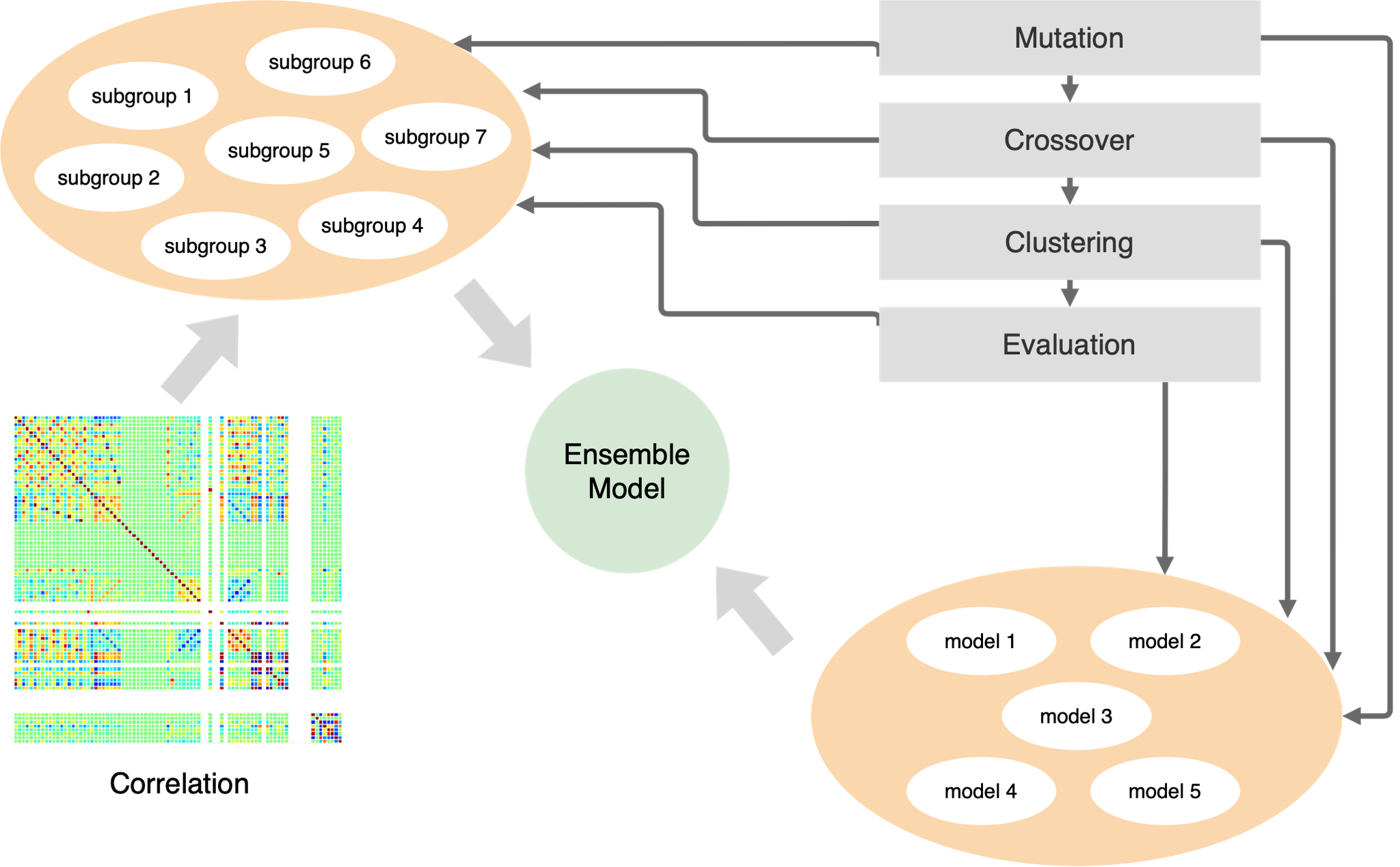}
    \caption{The architecture of the framework}
    \label{fig:neral_schema}
\end{figure*}

The prototype of our framework presented in figure \ref{fig:neral_schema} consists of two separate populations. The first is the models population. The second is the data subgroup population. This enables formation of the ensemble model using an approach that is similar to bagging-based technique. 
The framework starts with a generation of initial groups of features with the use of correlation (which is explained in detail in further subsections), then it mutates models and groups via the genetic algorithm. The final effect of those actions is an optimized ensemble model, that can be used to detect anomalies.

\subsection{General schema of the proposed solution}

During our experiments with various models, we noticed that almost all models detect a similar set of anomalies despite changes in their hyperparameters. Of course, the results were slightly different depending on the hyperparameters, but none of the changes had a significant impact on detection. Therefore, we decided to try to apply an ensemble model based on dividing available features into smaller groups and training each model on a separate subset of features. As a result of this, the models can discover more precise dependencies and relations between features. A simplified schema of our approach is presented in Algorithm \ref{alg:general}. We apply a neuroevolution approach for searching for an optimal partition into groups (line 1, algorithm \ref{alg:general}). After classifying data points using every model (line 2, algorithm \ref{alg:general}), we use a voting mechanism to determine whether a data point should be considered as an anomaly by the whole ensemble model (line 3, algorithm \ref{alg:general}).

\begin{algorithm}
\label{alg:general}
\SetAlgoLined
\KwResult{Classification of the anomalies}
Find the best partition of features into groups using a genetic algorithm\;
Train and evaluate a separate model for every group\;
Evaluate an ensemble model using a voting algorithm 
 \caption{Simplified general schema of our approach}
\end{algorithm}

To find an optimal partition of features into groups, we apply a genetic algorithm. The simplified schema of the genetic algorithms is presented in Algorithm \ref{alg:genetic}.

The single gene provides information that a feature $f$ is present in a group $t$. The single solution represents $k$ groups, each containing zero or more features. A sample solution for $k=3$ could be: $[[0, 1, 5, 12], [2, 3, 4, 9], [6, 7, 9]]$, where numbers in groups mean which features are present. Population $P$ contains $N_{P}$ solutions.

The parameters for the neuroevolution approach in this work are:
\begin{itemize}
    \item $k$ - maximal number of submodels in an ensemble model,
    \item $p_m$ - probability of the mutation in a single group of features,
    \item $N_{g}$ - number of generations in a genetic algorithm,
    \item $N_{P}$ - size of the population in a genetic algorithm,
    \item $N_{par}$ - number of parents mating,
    \item $N_{ep}$ - number of epochs to train while calculating fitness.
\end{itemize}

\begin{algorithm}
\label{alg:genetic}
\SetAlgoLined
\KwResult{Final population after $N_{g}$ generations}
\KwIn{$N_{P}$ - Size of population}
\KwIn{$N_{g}$ - Number of generations in the genetic algorithm}
\KwIn{$k$ - Max number of groups in a single solution}
Generate initial population \; %
$generation = 0$ \;
\While{$generation < N_{g}$} {
    For every solution in population calculate a fitness\;
    Choose the best solutions as a parents\;
    Create offspring using crossover\; 
    Mutate offspring\;
    $generation \gets generation + 1$
}
Return final population
 \caption{Genetic algorithm}
\end{algorithm}

\subsection{Elements of the genetic algorithm}

To improve convergence of a genetic algorithm, instead of using random initial population, we create it based on correlation between features. We use a hierarchical clustering with the addition of a little  randomness to achieve a diverse population.

The method for calculating fitness for a single solution is presented in Algorithm \ref{alg:fitness}. For every used dataset (SWAT and WADI), we split a normal part of the data into training and validation datasets. We calculate fitness for every feature group in the solution. As the first step, we train a chosen model on selected features from the training data for a given number of epochs (line 3, algorithm \ref{alg:fitness}). After that, we evaluate the trained model on training and validation data (lines 4 and 5, algorithm \ref{alg:fitness}) calculating losses. To normalise loss, we calculate the weighted loss from the training and validation datasets (lines 6 and 7) and we also divide weighted loss by the number of features in the group (line 8, algorithm \ref{alg:fitness}). The final fitness for every solution is calculated as a negated sum of losses for groups in the solution (lines 9 and 10, algorithm \ref{alg:fitness}). The value is negated because we want to minimise the total loss of an ensemble model, while in the genetic algorithm, the goal is to maximise the fitness. 

\begin{algorithm}
\label{alg:fitness}
\SetAlgoLined
\SetKwInOut{Input}{inputs}
\KwResult{Fitness value for solution $S$}
\KwIn{$S$ - solution}
\KwIn{$X_{t}$ - train dataset}
\KwIn{$X_{v}$ - validation dataset}
\KwIn{$N_{ep}$ - number of epochs to train while calculating fitness}
$loss_{sum} = [] $ \;
\For{$g$ $\in$ $S$} {
    $model = \TrainModel{(X_{t}, g, N_{ep})}$\;
    $loss_{t} = \EvaluateModel{(model, X_{t}, g)} $\;
    $loss_{v} = \EvaluateModel{(model, X_{v}, g)} $ \;
    $N_{X} = \Len{(X_{t})} + \Len{(X_{v})}$ \;
    $loss_{w} = \frac{\Len{(X_{t})}}{N_X} * loss_{t} + \frac{\Len{(X_{v})}}{N_X} * loss_{v}$\;
    $loss_{g} = \frac{loss_{w}}{\Len{(g)}}$\;
    $loss_{sum} = loss_{sum} + loss_{g}$
}
$return -loss_{sum} $
 \caption{Fitness calculation}
\end{algorithm}

During the crossover part (line 6, algorithm \ref{alg:genetic}), we create a new solution based on two selected parents. The detailed steps of the method are presented in Algorithm \ref{alg:crossover}. For every pair of groups of parents, we determine what range of features is present in the groups and choose random split point (lines 3-5, algorithm \ref{alg:crossover}). The new group for offspring then contains a parts of the groups from both parents (lines 6-13, algorithm \ref{alg:crossover}).

\begin{algorithm}
\label{alg:crossover}
\SetAlgoLined
\SetKwInOut{Input}{inputs}
\KwResult{$S_{new}$ - new solution after crossover}
\KwIn{$S1$ - parent solution}
\KwIn{$S2$ - parent solution}
$S_{new} = [] $ \;
\For{$g1, g2$ $\in$ $zip(S1, S2)$} {
  $min_{g1g2} = \Min{(\Min{(g1)}, \Min{(g2)})}$ \;
  $max_{g1g2} = \Max{(\Max{(g1)}, \Max{(g2)})}$ \;
  $split\_point = \Randint{(min_{g1g2}, max_{g1g2})}$\;
  $g_{new} = [] $ \;
  \For{$f$ $\in$ $g1$} {
    \If{$f < split\_point$} {
      $g_{new}.add(f)$ \;
    }
  }
  \For{$f$ $\in$ $g2$} {
    \If{$f > split\_point$} {
      $g_{new}.add(f)$ \;
    }
  }
  $S_{new}.add(g_{new})$
}
$\ReturnV{S_{new}}$
 \caption{Crossover algorithm}
\end{algorithm}

Offspring created via a crossover algorithm can also be affected by mutations (line 7, algorithm \ref{alg:genetic}). In our work, we use three types of mutation:
\begin{enumerate}
    \item Duplicating selected feature to other group (presented in Algorithm \ref{alg:moving_mutation}),
    \item Vanishing features that exists in more than one group in a single solution (presented in Algorithm \ref{alg:vanishing_mutation}),
    \item Adding features that do not exist in any group in a single solution (presented in Algorithm \ref{alg:new_features_mutation}).
\end{enumerate}
The goal of mutations is to help to maintain diversity in the population. Mutation 1 allows having the same feature available in a few groups. Mutation 2 protects solutions from having a few groups with exactly the same features and from overusing any feature. Mutation 3 makes it possible to restore features lost in other genetic operations.

\begin{algorithm}
\label{alg:moving_mutation}
\SetAlgoLined
\SetKwInOut{Input}{inputs}
\KwResult{Mutated solution}
\KwIn{$S$ - solution }
\KwIn{$P_m$ - probability of a mutation}
\For{$i = 0;\ i < k;\ i = i + 1$}{
  \If{$\RandomRandom{()}  < P_m$}{
   $feature\_to\_move = \RandomChoice{(S[i])} $ \;
   $n_{next\_group} = (j+1) \bmod k $
   $S[n_{next\_group}].add(feature\_to\_move) $\;
   }
 }
 $\ReturnV{S}$
 \caption{Moving mutation}
\end{algorithm}

\begin{algorithm}
\label{alg:vanishing_mutation}

\SetAlgoLined
\SetKwInOut{Input}{inputs}
\KwResult{Mutated solution}
\KwIn{$S$ - solution}
\KwIn{$P_m$ - probability of a mutation}
$count_{f} = \DefaultDict{(0)}$ \;
\For{$group$ $\in$ $S$} {
  \For{$feature$ $\in$ $group$} {
    $count_{f}[feature] += 1$ \;
  }
}

\For{$group$ $\in$ $S$} {  
  \For{$feature$ $\in$ $group$} {
    \If{$\RandomRandom{()} > \frac{1}{count_{f}[i]}$}{
        $group.remove(feature)$ \;
    }
  }
}

return $S$
\caption{Vanishing mutation}
\end{algorithm}

\begin{algorithm}
\label{alg:new_features_mutation}
\SetAlgoLined
\SetKwInOut{Input}{inputs}
\KwResult{Mutated solution}
\KwIn{$S$ - solution}
\KwIn{$FS$ - features space}
\For{$feature$ $\in$ $FS$} {
  \If{$feature$ $ \notin S$} {
      \For{$group$ $\in$ $S$} {
        \If{$\RandomRandom{()} > \frac{1}{k}$}{
            $group.add(feature)$ \;
        }
      }
  }
}
\Return{$S$}
 \caption{New features mutation}
 \centering
\end{algorithm}

\section{Results}
In this section, we describe used datasets and models. We also demonstrate the improvements that were possible to achieve by the usage of proposed solutions. We provide a comparison with methods from the state of the work articles. All of our presented calculations were performed on the Nvidia Tesla V100-SXM2-32GB\footnote{https://www.nvidia.com/en-us/data-center/v100/}. In order to reduce both training times i.e during the evolution algorithm and the final training, each subgroup is calculated on the separate GPGPU. The values of the parameters of the genetic algorithm were the same for all experiments and are presented in Table \ref{tab:genetic_params} (column \textit{basic value}). Moreover, the model which gained the best results (CNN 1D) was also run once again with a higher value of the following parameters: \textit{population size} and \textit{parents mating} (column \textit{Rerun value} in table \ref{tab:genetic_params}) to check how it would affect the efficiency of the algorithms.

\begin{table}[]
\centering
\caption{Parameters of genetic algorithm}
\label{tab:genetic_params}
\begin{tabular}{|l|c|c|}
\hline
\textbf{Parameter}                & \textbf{Basic value} &  \textbf{Rerun Value} \\ \hline
Population size          & 8  &  16 \\ \hline
Number of parents mating & 4   &  8\\ \hline
Mutation probability     & 0.1  & 0.1\\ \hline
Number of generations    & 10   & 10 \\ \hline
\end{tabular}
\end{table}

\subsection{Datasets}
\begin{table}[]
\centering
\caption{Statistics of the used datasets}
\begin{tabular}{|c|c|c|c|c|}
\hline
\textbf{Datasets} & \textbf{\#Features} & \textbf{\#Train} & \textbf{\#Test} & \textbf{\#Anomalies} \\ \hline
\textbf{SWAT}     & 51                  & 49668            & 44981           & 11.97\%              \\ \hline
\textbf{WADI-2017}     & 123                 & 1048571           & 172801          & 5.99\%               \\ \hline
\textbf{WADI-2019}     & 123                 & 784571           & 172801          & 5.77\%               \\ \hline
\end{tabular}
\label{statistc_of_datasets}
\end{table}
As training and testing data, the following were used: 
\begin{itemize}
    \item \textit{Secure Water Treatment (SWaT) Dataset} \cite{SWAT} - it contains data gathered from a scaled-down version of a real water treatment plant. Data were collected for an 11-day period in two modes - 7 days of a normal operation of the plant and 4 days during which there were cyber and physical attacks executed.
    \item \textit{Water Distribution (WADI) Dataset} \cite{WADI} - this dataset contains data from a scaled-down version of a water distribution network in a city. Collected data contains 14 days of normal operation and 2 days during which there were 15 attacks executed. As present in Table \ref{statistc_of_datasets}, there are two WADI collections from 2017 and 2019 available. In our experiments, we are using the newest version as this is recommended by the authors of the dataset.
\end{itemize}

\subsection{Models}

In this paper, in order to detect anomaly we are using three models of the autoencoder. The first of these was proposed in \cite{cnn_auto-encoder} where the encoder contains three 1D CNN layers with kernel sizes $k_1=8$, $k_2=6$, $k_3=4$ and filter maps $f_1=64$, $f_2=128$, $f_3=256$. Each CNN layer is followed by LReLU \cite{Maas13rectifiernonlinearities} activation function and batch normalisation calculation. The decoder is a mirror reflection of the encoder where CNN layers are replaced by transposed CNN layers. The second model is a variational autoencoder \cite{lstm-vae} where both the encoder and the decoder contain two LSTM layers with hidden sizes equalling 16 which are followed by the LReLU activation function. In the each case of the training phase the batch size is set at 32 and over the test phase, it is set at 1. The third model is the USAD model proposed in the literature \cite{USAD}. It utilises the idea of GAN networks and the architecture of autoencoders. The USAD model consist of two autoencoders built from one shared encoder and two decoders. They are trained with the use of proposed two-phase training which includes standard autoencoder training and adversarial training specific to GAN networks.

The architectures of the models were partially explored by us, including searching for optimal window size, number of layers, and numbers of neurons in layers. We use the found hyperparameters in the neuroevolution approach. A more advanced search is planned for future works (see section \ref{section:future_work}). In the evolution algorithm (see Section \ref{neuro-evolution}), each model is trained over 15 epochs, and during the final training, each model is trained over 70 epochs. We divide the multivariate time series into sub-sequences with a sliding window and we determine the size of this parameter in an experimental way. Consequently, sliding windows have sizes of 4 for autoencoders with CNN layers, 8 in the case of LSTM-VAE and 12 in the case of USAD. In order to speed up training, we are using down-sampling with the ratio 5, which reduces the size of the data. As was indicated in the literature \cite{USAD} this operation did not cause a significant drop in accuracy. Table \ref{parameters_of_model} contains the number of trainable parameters for each model used, the optimal values for some of them and the time that is necessary to perform the whole process. As it turns out, the most parameters are included in the USAD model and therefore this takes the longest time to train. 

\begin{table}[]
\caption{Parameters of the models}
\centering
\label{parameters_of_model}
\begin{tabular}{|l|c|c|c|}
\hline
\multicolumn{1}{|c|}{\textbf{Method}} & \textbf{\begin{tabular}[c]{@{}c@{}}Neuroevolution \\ time\end{tabular}} & \textbf{\begin{tabular}[c]{@{}c@{}}\#Trainable \\ Parameters\end{tabular}} & \textbf{Type of parameters}                                                                                                                  \\ \hline
\textbf{LSTM-VAE}                     & 24h                                                               & 2 378 496                                                                    & \begin{tabular}[c]{@{}c@{}}sliding widow=4\\ \#final training epochs=70\\ \#epochs during fitness=15\end{tabular}                            \\ \hline
\textbf{USAD}                         & 32h                                                               & 3 937 360                                                                     & \begin{tabular}[c]{@{}c@{}}sliding windows = 12\\ \#final training epochs=70\\ \#epochs during fitness=15\end{tabular}                        \\ \hline
\textbf{CNN 1D}                       & 16h                                                               & 366 476                                                                     & \begin{tabular}[c]{@{}c@{}}sliding window=4,\\ \#final training epochs=70,\\ \#epochs during fitness=15,\\ learning rate = 0.01\end{tabular} \\ \hline
\end{tabular}
\end{table}

\subsection{Experiments}
Table \ref{result_without_groups} contains collected results from \cite{PCA}\cite{KNN}\cite{DAGMAM}\cite{lstm-vae} \cite{USAD}\cite{MAD-GAN}\cite{graph-nn}. Those results were achieved based on SWAT and WADI-2017 dataset. Additionally, this table contains results which were generated for this paper and in the case of WADI dataset, wadi 2019 were used (marked as *).  Moreover, for our baseline models (USAD, LSTM-VAE, CNN 1D), we present outcomes from our experiments for the SWAT dataset, in which for some cases, the results are slight different to the original result, as it required preparing our own implementation (marked as **). 

Table \ref{tab:result_with_group} contains gained results after introducing splitting into the groups through the use of the genetic algorithms. We can observe a huge improvement on the WADI dataset in the case of the USAD model and CNN-based autoencoder. It has the smallest impact in the LSTM-VAE model. 
The best results were gained for the CNN 1D autoencoder. Due to that, for the CNN 1D model, we rerun the experiment with a higher value of the following parameters of the genetic algorithm: \textit{population size} and \textit{parents mating}. As a result, it was possible to improve the F1-score by about 2\% in the case of SWAT and WADI datasets. These results are marked as (***) in the table \ref{tab:result_with_group}.

\begin{table}[]
\centering
\caption{Anomaly detection accuracy (precision (\%), recall(\%), F1-score(\%)) on two datasets without splinting into groups. Results marked as * was generated by usage of the WADI-2019 dataset. ** means that we had to reimplement a model on our own}
\begin{tabular}{llll|lll}
\hline
                  & \multicolumn{3}{c|}{\textbf{SWAT}}         & \multicolumn{3}{c}{\textbf{WADI}}          \\ \cline{2-7} 
\textbf{Method}   & \textbf{Prec} & \textbf{Rec} & \textbf{F1} & \textbf{Prec} & \textbf{Rec} & \textbf{F1} \\ \hline
\textbf{PCA}      & 24.92         & 21.63        & 0.23        & 39.53         & 5.63         & 0.10        \\
\textbf{KNN}      & 7.83          & 7.83         & 0.08        & 7.76          & 7.75         & 0.08        \\
\textbf{DAGMAM}   & 27.46         & 69.52        & 0.39        & 54.44         & 26.99        & 0.36        \\
\textbf{LSTM-VAE} & 96.24         & 59.91        & 0.74        & 87.79         & 14.45        & 0.25        \\
\textbf{MAD-GAN}  & 98.97         & 63.74        & 0.77        & 41.44         & 33.92        & 0.37        \\
\textbf{USAD}     & 98.51         & 66.18        & 0.79        & 99.47         & 13.18        & 0.23        \\
\textbf{USAD**}   & 88.21         & 65.29        & 0.75        & 26.28*         & 35.31*        & 0.30*        \\
\textbf{CNN 1D}   & 94.25         & 67.92        & 0.78        & 39.30*         & 20.28*        & 0.27*        \\
\textbf{GDN}      & 99.35         & 68.12        & 0.81        & 97.50         & 40.19        & 0.57       
\end{tabular}
\label{result_without_groups}
\end{table}

\begin{table}[]

\caption{Anomaly detection accuracy (precision (\%), recall(\%), F1-score(\%)) on two datasets after splitting into groups}
\centering
\begin{tabular}{llll|lll}
\hline
                  & \multicolumn{3}{c|}{\textbf{SWAT}}         & \multicolumn{3}{c}{\textbf{WADI*}}          \\ \cline{2-7} 
\textbf{Method}   & \textbf{Prec} & \textbf{Rec} & \textbf{F1} & \textbf{Prec} & \textbf{Rec} & \textbf{F1} \\
\textbf{LSTM-VAE} & 95.69             & 55.18            & 0.72           &21.22             & 29.12            & 0.28           \\
\textbf{USAD}     & 98.10         & 66.01        & 0.79        & 71.24             & 31.41            & 0.43           \\
\textbf{CNN 1D}   & 95.24         & 63.73        & 0.78        & 63.76         & 43.54        & 0.52  \\

\textbf{CNN 1D***}   & 93.61         & 69.40        & 0.80        & 79.35         & 41.23        & 0.54  
\end{tabular}
\label{tab:result_with_group}
\end{table}

\section{Conclusions}

The results show that data distribution and dividing the input signals to subgroups and a feeding ensemble model can significantly improve the efficiency of the anomaly detection process. The neuro-evolution process helps to find near optimal subgroups. The tests were run on WADI and SWAT benchmarks. In both cases best results were achieved 
among no graph neural network models. The improvements on WADI dataset are significant. The reason of this fact is 
more sensors and time series samples than in SWAT dataset.

\section{Future work}
\label{section:future_work}
The paper presents a framework for evolving ensemble deep learning autoencoders for anomaly detection. Future work will concentrate on further enhancements of the algorithm. The most important enhancements are ensemble model based on graph networks, new crossover to mix different architectures together e.g. attention with discriminator, and graph networks with USAD. The main work will concentrate on evolving optimal autoencoder architecture. The last action would be to run longer simulations which can give further improvements in F1 score. Further simulations will be run in bigger populations and with more iterations.

\section*{Acknowledgment}
\textit{This research was supported in part by PLGrid Infrastructure.}

\nocite{*} 
\bibliographystyle{unsrt}
\bibliography{bibliograph}

\begin{thebibliography}{10}

\bibitem{graph-nn}
Ailin Deng and Bryan Hooi.
\newblock Graph neural network-based anomaly detection in multivariate time
  series, 2021.

\bibitem{PCA}
Rolf Isermann.
\newblock Model-based fault detection and diagnosis - status and applications.
\newblock {\em IFAC Proceedings Volumes}, 37(6):49--60, 2004.
\newblock 16th IFAC Symposium on Automatic Control in Aerospace 2004,
  Saint-Petersburg, Russia, 14-18 June 2004.

\bibitem{KNN}
Fabrizio Angiulli and Clara Pizzuti.
\newblock Fast outlier detection in high dimensional spaces.
\newblock In Tapio Elomaa, Heikki Mannila, and Hannu Toivonen, editors, {\em
  Principles of Data Mining and Knowledge Discovery}, pages 15--27, Berlin,
  Heidelberg, 2002. Springer Berlin Heidelberg.

\bibitem{LOF}
Markus~M. Breunig, Hans-Peter Kriegel, Raymond~T. Ng, and J\"{o}rg Sander.
\newblock Lof: Identifying density-based local outliers.
\newblock {\em SIGMOD Rec.}, 29(2):93–104, May 2000.

\bibitem{one-class-svm}
J.~Ma and S.~Perkins.
\newblock Time-series novelty detection using one-class support vector
  machines.
\newblock In {\em Proceedings of the International Joint Conference on Neural
  Networks, 2003.}, volume~3, pages 1741--1745 vol.3, 2003.

\bibitem{Isolation_forest}
Fei~Tony Liu, Kai~Ming Ting, and Zhi-Hua Zhou.
\newblock Isolation forest.
\newblock In {\em 2008 Eighth IEEE International Conference on Data Mining},
  pages 413--422, 2008.

\bibitem{DAGMAM}
Bo~Zong, Qi~Song, Martin~Renqiang Min, Wei Cheng, C.~Lumezanu, Dae ki~Cho, and
  H.~Chen.
\newblock Deep autoencoding gaussian mixture model for unsupervised anomaly
  detection.
\newblock In {\em ICLR}, 2018.

\bibitem{lstm-vae}
Daehyung Park, Yuuna Hoshi, and Charles~C. Kemp.
\newblock A multimodal anomaly detector for robot-assisted feeding using an
  lstm-based variational autoencoder.
\newblock {\em IEEE Robotics and Automation Letters}, 3(3):1544--1551, 2018.

\bibitem{cnn_auto-encoder}
Stefania Russo, Andy Disch, Frank Blumensaat, and Kris Villez.
\newblock Anomaly detection using deep autoencoders for in-situ wastewater
  systems monitoring data, 2020.

\bibitem{USAD}
Julien Audibert, Pietro Michiardi, Fr\'{e}d\'{e}ric Guyard, S\'{e}bastien
  Marti, and Maria~A. Zuluaga.
\newblock Usad: Unsupervised anomaly detection on multivariate time series.
\newblock In {\em Proceedings of the 26th ACM SIGKDD International Conference
  on Knowledge Discovery Data Mining}, KDD '20, page 3395–3404, New York, NY,
  USA, 2020. Association for Computing Machinery.

\bibitem{MAD-GAN}
Dan Li, Dacheng Chen, Baihong Jin, Lei Shi, Jonathan Goh, and See-Kiong Ng.
\newblock Mad-gan: Multivariate anomaly detection for time series data with
  generative adversarial networks.
\newblock In Igor~V. Tetko, V{\v{e}}ra K{\r{u}}rkov{\'a}, Pavel Karpov, and
  Fabian Theis, editors, {\em Artificial Neural Networks and Machine Learning
  -- ICANN 2019: Text and Time Series}, pages 703--716, Cham, 2019. Springer
  International Publishing.

\bibitem{neuroevol_overview}
P.~Mooney E.~Galvan.
\newblock Neuroevolution in deep neural networks: Current trends and future
  challenges.
\newblock {\em CoRR abs/2006.05415}, Jun 2020.

\bibitem{sceneNet}
A.~Ma, Y.~Wan, Y.~Zhong, and J.~Wang.
\newblock Scenenet: Remote sensing scene classification deep learning network
  using multi-objective neural evolution architecture search.
\newblock {\em ISPRS Journal of Photogrammetry and Remote Sensing pp. 171-188},
  DOI:10.1016/j.isprsjprs.2020.11.025, February 2021.

\bibitem{neuro_image}
Yanan Sun, Bing Xue, Mengjie Zhang, and Gary~G. Yen.
\newblock Evolving deep convolutional neural networks for image classification.
\newblock {\em CoRR abs/1710.10741}, Oct 2017.

\bibitem{neuro_evolving}
R.~Miikkulainen, J.~Liang, E.~Meyerson, A.~Rawal, D.~Fink, O.~Francon, B.~Raju,
  H.~Shahrzad, A.~Navruzyan, N.~Duffy, and B.~Hodjat.
\newblock Evolving deep neural networks.
\newblock {\em CoRR abs/1703.00548}, Mar 2017.

\bibitem{Mirsky2018KitsuneAE}
Yisroel Mirsky, Tomer Doitshman, Y.~Elovici, and Asaf Shabtai.
\newblock Kitsune: An ensemble of autoencoders for online network intrusion
  detection.
\newblock {\em ArXiv}, abs/1802.09089, 2018.

\bibitem{2020MalwareDetection}
Xiang Jin, Xiaofei Xing, Haroon Elahi, Guojun Wang, and Hai Jiang.
\newblock A malware detection approach using malware images and autoencoders.
\newblock In {\em 2020 IEEE 17th International Conference on Mobile Ad Hoc and
  Sensor Systems (MASS)}, pages 1--6, 2020.

\bibitem{2018StreamLearning}
Yue Dong and Nathalie Japkowicz.
\newblock Threaded ensembles of autoencoders for stream learning.
\newblock {\em Computational Intelligence}, 34(1):261--281, 2018.

\bibitem{SWAT}
Aditya~P. Mathur and Nils~Ole Tippenhauer.
\newblock Swat: a water treatment testbed for research and training on ics
  security.
\newblock In {\em 2016 International Workshop on Cyber-physical Systems for
  Smart Water Networks (CySWater)}, pages 31--36, 2016.

\bibitem{WADI}
Chuadhry Ahmed, Venkata Palleti, and Aditya Mathur.
\newblock Wadi: a water distribution testbed for research in the design of
  secure cyber physical systems.
\newblock pages 25--28, 04 2017.

\bibitem{Maas13rectifiernonlinearities}
Andrew~L. Maas, Awni~Y. Hannun, and Andrew~Y. Ng.
\newblock Rectifier nonlinearities improve neural network acoustic models.
\newblock In {\em in ICML Workshop on Deep Learning for Audio, Speech and
  Language Processing}, 2013.

\bibitem{NEAT}
Kenneth~O. Stanley and Risto Miikkulainen.
\newblock Evolving neural networks through augmenting topologies.
\newblock {\em Evolutionary Computation}, 10(2):99--127, 2002.

\end{thebibliography}
\end{document}